\documentclass[conference]{IEEEtran}
\usepackage{cite}
\usepackage{amsmath,amssymb,amsfonts}
\usepackage{algorithmic}
\usepackage{graphicx}
\usepackage{textcomp}
\usepackage{xcolor}
\graphicspath{{figs/}}
\usepackage{comment}

\usepackage{spconf}
\usepackage{caption}
\usepackage{subcaption}
\DeclareMathOperator*{\argmax}{arg\, max}
\DeclareMathOperator*{\argmin}{arg\, min}
\def\BibTeX{{\rm B\kern-.05em{\sc i\kern-.025em b}\kern-.08em
    T\kern-.1667em\lower.7ex\hbox{E}\kern-.125emX}}
    
\usepackage[nomargin,inline,final]{fixme}  
\FXRegisterAuthor{pf}{apf}{{\color{blue}pf}}
    
\begin{document}

\title{Graph Heat Mixture Model Learning}

\name{Hermina Petric Maretic \qquad Mireille El Gheche \qquad Pascal Frossard}

\address{Ecole Polytechnique F\'ed\'erale de Lausanne (EPFL), Signal Processing Laboratory (LTS4)}

\maketitle

\begin{abstract}
Graph inference methods have recently attracted a great interest from the scientific community, due to the large value they bring in data interpretation and analysis. However, most of the available state-of-the-art methods focus on scenarios where all available data can be explained through the same graph, or groups corresponding to each graph are known \textit{a priori}. In this paper, we argue that this is not always realistic and we introduce a generative model for mixed signals following a heat diffusion process on multiple graphs. We propose an expectation-maximisation algorithm that can successfully separate signals into corresponding groups, and infer multiple graphs that govern their behaviour. We demonstrate the benefits of our method on both synthetic and real data. 
\end{abstract}


\section{Introduction}

Understanding pairwise relationships is often crucial in interpreting and analysing high-dimensional data. While these relationships are sometimes given explicitly in the dataset (e.g., data from social, biological or sensor networks), many datasets do not have a readily available graph structure modelling relationships between data.
Network inference deals precisely with such data, providing means to better represent, understand and eventually analyze data. 

First efforts in inferring data relationships came in terms of sparse inverse covariance (precision) matrix inference \cite{dempster1972covariance}, where pairwise relationships are modelled as conditional dependencies between nodes. More recent works focus on structured graph representations, such as (generalised) graph Laplacian matrices\cite{dong2018learning}. A standard assumption is that of signal smoothness that permits to develop learning algorithms with a signal processing perspective \cite{Dong14,kalofolias2016learn, egilmez2017graph}. Another path of works assume that the data are generated by a heat diffusion process on an unknown graph. This is a commonly used model, with applications stemming from brain diffusion modelling to social networks \cite{ma2008mining}. Several works have studied the graph inference problem from heat diffusion signals, including sparse dictionary models \cite{thanou2017learning}, online graph inference \cite{vlaski2018online} and models that can deal with more general diffusion signals \cite{egilmez2018graph, segarra2017network, maretic2017graph}. Some works have considered data which is \textit{a priori} grouped into multiple clusters, and each of these clusters can be represented with a different graph \cite{kalofolias2017learning} \cite{segarra2017joint}. However, it is not always reasonable to assume clusters are predefined or easily obtainable.
 
In this work, we build on our prior work on the Graph Laplacian mixture model \cite{maretic2018graph} and propose a generative model for mixed signals that follow a heat diffusion process on different graphs. Specifically, each signal belongs to a cluster and follows a heat diffusion process on a graph corresponding to its cluster. However, both the clusters and the graphs are assumed to be unknown. We present a novel algorithm that can jointly separate signals into clusters that relate to the generative graphs, and efficiently infer the corresponding graph structures. The algorithm relies on a well established expectation maximisation scheme, while the graph learning step is formulated in a convex manner and can be efficiently solved with FISTA \cite{beck2009fast}. We compare our method to existing works that take into account a simple smoothness assumption \cite{maretic2018graph} or implicitly learn graph structures, as well as a separated clustering and graph inference scheme. We show the benefits of our model in terms of both signal clustering and multiple graph inference on synthetic data and real data describing Uber pick-ups in New York city. 

This is one of the first methods for multiple graph inference from mixed signals. We believe this is an important area that brings a new dimension to graph inference and hope our method will provide valuable insights in many complex datasets.
\section{Preliminaries}

Let $\{\mathcal{G}_{k} = (V,E_k, W_k)\}_{1\leq k \leq K}$ be a collection of undirected, weighted graphs with a set of $N$ shared vertices $V$. Each graph $\mathcal{G}_k$ has a separate set of edges $E_k$, while $W_k = [w_k^{i,j}]_{i,j} \in \mathbb{R}^{N\times N}$ are the weighted adjacency matrices, with $w_k^{i,j} \geq 0$, and $w_k^{i,i} = 0$ for all $i,j$. 

The Laplacian matrix of $\mathcal{G}_k$ is defined as
\begin{equation}
L_k = D_k - W_k,
\end{equation}
where $D_k$ is a diagonal matrix of node degrees.
A signal $x\in \mathbb{R}^N$ that lives on the graph $\mathcal{G}_k$ and corresponds to a heat diffusion process, is defined as \cite{chung2007heat}:
\begin{align}
    x = e^{-\tau L_k} w 
\end{align}
where $\tau$ is the heat diffusion parameter. Throughout this paper, we will assume $w \sim N(0,I)$.


\section{Graph heat mixture model}
\begin{figure}[t]
	\centering
	\includegraphics[scale = 0.13]{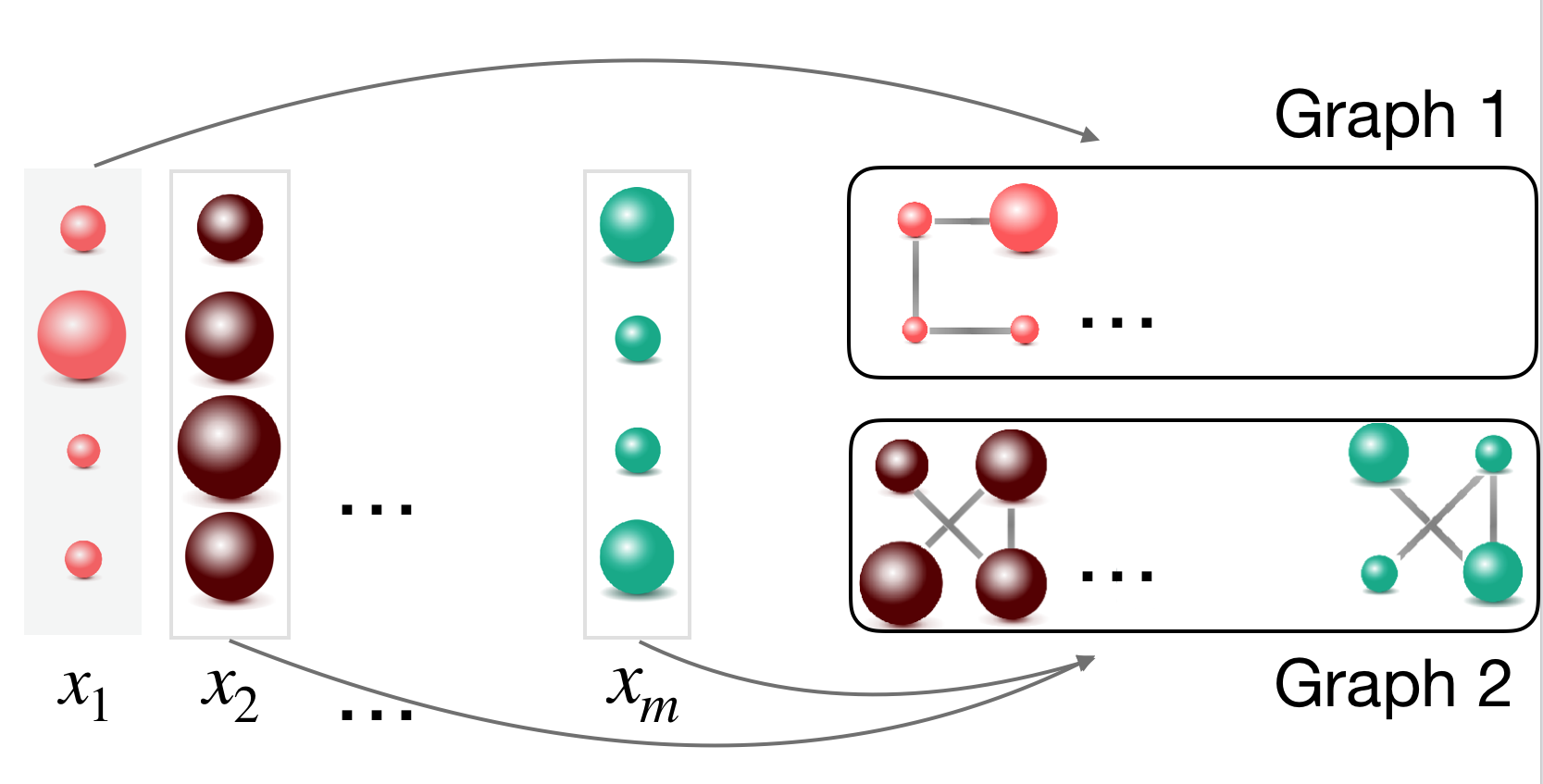}
	\caption{A toy example for the graph heat mixture model. The signals $x_1,x_2, \cdots, x_m$ live each on exactly one of the two proposed graphs, $\textup{Graph\;1}$ and $\textup{Graph\;2}$. Our objective is to separate the signals into clusters corresponding to each graph, while inferring both graph structures at the same time.}\label{fig:glmm_drawing}
\end{figure}

We consider a set of observed signals $X=[x_1 \;|\; \dots \;|\; x_M] \in \mathbb{R}^{N\times M}$, where each signal $x_m$ is associated with one of the graphs $\mathcal{G}_{k}$, for every $m\in \{1,\cdots,M\}$ and $k \in \{1,\cdots,K\}$. As shown in Figure \ref{fig:glmm_drawing}, the set of signals associated to the same graph $\mathcal{G}_{k}$ defines a cluster $C_k$. 

\begin{figure}[h]
	\centering
	\includegraphics[scale = 0.3]{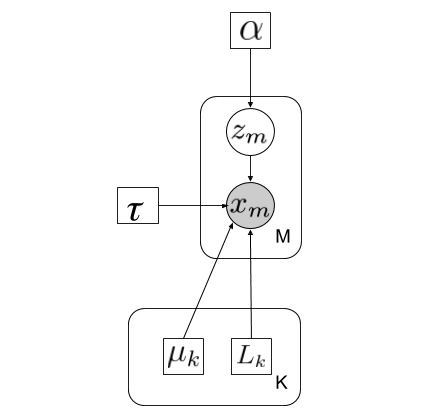}
	\caption{Plate notation for our generative model. Filled in circles are observed variables, small empty squares are unknown parameters, and non-filled circles represent latent variables. Large plates indicate repeated variables.}\label{fig:ghmm_plate}
\end{figure}

We propose a generative model for such signals. Each signal $x_m$ is associated to cluster $C_k$ with probability $\alpha_k$. As shown in Figure \ref{fig:ghmm_plate}, this selection is modelled through a latent variable $z_m$, such that 
\begin{equation}
z_{m} (k) =\left\{
\begin{array}{@{}ll@{}}
1, & \text{if}\ x_m \in C_k \\
0, & \text{otherwise.}
\end{array}\right.
\end{equation}
This directly defines a prior probability for the latent variable $\boldsymbol z$, where $p(z_{m}(k)=1) = \alpha_k$.
Further, the signals in cluster $C_k$ share a mean $\mu_k$ and follow a heat diffusion process on graph $L_k$, yielding:
\begin{align}
\label{eq:heatdiff}
x_m |(z_{m} (k) = 1)= \mu_k + e^{-\tau L_k}w_m, \quad w_m\sim N(0, I).
\end{align}
We can now model the probability distribution of $x_m$ as
\begin{align}
x_m| (z_{m} (k) = 1) \sim N (\mu_k, e^{-2\tau L_k}).
\end{align}
Marginalising over all possible clusters $C_k$, we have:
\begin{align}
p(x_m) &= \sum_{k=1}^K p(z_{m}(k)=1) p(x_m| z_{m}(k) = 1) \nonumber \\
&= \sum_{k=1}^K \alpha_k N (\mu_k, e^{-2\tau L_k}).
\end{align}
Finally, taking all $M$ independent signals  into account, the probability distribution for $X$ becomes:
\begin{align}
\label{eq:model}
p(X) = \prod_{m=1}^M \sum_{k=1}^K \alpha_k N (\mu_k, e^{-2\tau L_k}).
\end{align}

\subsection{Problem formulation}

Given the model in Eq. \eqref{eq:model}, we want to infer the unknown clusters and graph structures from the observed signals $X$. Specifically, we formulate a maximum likelihood estimation problem:
\begin{align}
\label{eq:likelihood}
\argmax_{\boldsymbol\alpha, \boldsymbol\mu, \boldsymbol L, \tau} \enskip & \text{ln } \prod_{m=1}^M \sum_{k=1}^K \alpha_k N (\mu_k, e^{-2\tau L_k})=  \nonumber \\  
	\argmax_{\boldsymbol\alpha, \boldsymbol\mu, \boldsymbol L, \tau} \enskip  &\sum_{m=1}^{M} \text{ln }  \sum_{k=1}^K \alpha_k N(x_m|\mu_k, e^{-2\tau L_k}),
\end{align}
where the optimisation variable $\boldsymbol\alpha$ relates to cluster allocation, the variable $\boldsymbol L$ represents the graphs, and the variables $\boldsymbol\mu$ and $\tau$ characterize the heat diffusion processes. The optimisation problem in Eq. \eqref{eq:likelihood} is very difficult to solve directly, and we propose to solve it with an expectation-maximisation (EM) algorithm in the next Section. 

\section{Inference Algorithm}

We propose here to solve the inference problem of Eq. \eqref{eq:likelihood} with an alternating EM algorithm, as it is commonly done for problems of the same form. We randomly initialise the values $\boldsymbol\alpha$, $\boldsymbol\mu$ and $\boldsymbol L$. Then, we alternate between an expectation step, where we estimate expected values for latent variables $\boldsymbol{z}$ and a maximisation step, where we use these expected values to uncover unknown parameters $\boldsymbol\alpha$, $\boldsymbol\mu$ and $\boldsymbol L$. As will be shown later, $\tau$ acts as a scaling factor for $\boldsymbol L$, and only has a unique solution if there is some additional knowledge about $\boldsymbol L$. The two steps of the algorithm are described in more details below.

In the expectation step of the algorithm we estimate cluster responsibilities $\gamma_{m}(k)$. They are the expected values of latent variables $z_{m}(k)$ and the best estimation for the clusters of $x_m$ given the observed data and the current version of parameters $\boldsymbol\alpha$, $\boldsymbol\mu$ and $\boldsymbol L$. Formally, these cluster responsibilities can be estimated as:
\begin{align}
\gamma_{m}(k) &= p(z_{m}(k) = 1 |x_m, \mu_k, L_k) \nonumber \\
&= \frac{p(z_{m}(k) = 1) p(x_m| z_{m}(k) = 1, \mu_k, L_k)}{\sum_{l=1}^K p(z_{m} (l) = 1) p(x_m| z_{m}(l) = 1, \mu_l, L_l)} \nonumber \\
&= \frac{\alpha_k N(x_m|\mu_k,e^{-2\tau L_k})}{\sum_{l=1}^K \alpha_l N(x_m|\mu_l,e^{-2\tau L_l})}.
\end{align}
This closed form solution permits to infer the entire matrix of cluster responsibilities $\boldsymbol{\gamma} \in R^{M\times K}$.

With the estimated responsibilities $\gamma_{m}(k)$, we can move to the maximisation step. Specifically, we maximise the optimisation problem of Eq. \eqref{eq:likelihood} over the expected posterior distribution given all observations (for details, see \cite{maretic2018graph}): 
\begin{align}
\argmax_{\boldsymbol \alpha, \boldsymbol\mu,\boldsymbol L, \tau} \enskip   &\sum_{m=1}^{M} \sum_{k=1}^K \gamma_{m}(k) \text{log} (\alpha_k N(x_m|\mu_k, e^{-2\tau L_k})) = \nonumber \\
 \argmax_{\boldsymbol \alpha, \boldsymbol\mu,\boldsymbol L, \tau} \enskip   &\sum_{m=1}^{M} \sum_{k=1}^K \gamma_{m}(k) (\text{log} \alpha_k + \text{log} N(x_m|\mu_k, e^{-2\tau L_k})). \label{eq:map_2}
\end{align}
It is not difficult to infer a closed form solution for $\boldsymbol\alpha$ and $\boldsymbol\mu$, with:
\begin{align} 
\alpha_k &= \frac{\sum_{m=1}^{M} \gamma_{m}(k)}{M }. \label{eq:alpha}\\
\mu_k &= \frac{\sum_{m=1}^{M}\gamma_{m}(k) x_m}{\sum_{m=1}^{M} \gamma_{m}(k) }. \label{eq:mu}
\end{align}
To infer graph Laplacian matrices $\boldsymbol L$, we first notice that the covariance matrices $\Sigma_k = e^{-2\tau L_k}$ that relates to the heat diffusion processes, can also be written in closed form as:
\begin{align}
\Sigma_k : = \frac{\sum_m \gamma_{m}(k)(x_m - \mu_k)(x_m - \mu_k)^T}{\sum_m \gamma_{m}(k)}
\end{align}
In order to efficiently infer graph structures, the information of data probability might however not be sufficient. Namely, without very large amounts of data, the sample covariance matrices $\{\Sigma_k\}$ are usually noisy (if not low rank), and it can be difficult to recover the exact structure of the graph matrix. We thus formulate a problem that aims at finding a valid Laplacian matrix that would give a covariance matrix similar to the sample covariance one, while at the same time imposing a graph sparsity constraint. Namely, we can estimate the weight matrix $W_k$ as
\begin{align}
& \qquad \argmin_{W_k} \|\Sigma_k - e^{-2\tau L_k}\|_F^2 + \beta \|W_k\|_1, \nonumber \\
\rm{s.t.} & \left\{\begin{aligned}  & L_k = D_k - W_k \\ & 
\mathcal{W} = \{W_k \in \mathbb{R}_+^{N\times N} : W_k = W_k^T, diag(W_k) = 0\} 
\end{aligned}\right.
\end{align}
This is equivalent to solving 
\begin{align}
& \argmin_{W_k \in \mathcal{W}} \|\text{log} \Sigma_k + 2\tau L_k\|_F^2 + \beta \|W_k\|_1 
\end{align}
with the same constraints. It results in a convex problem that can be solved efficiently with FISTA \cite{beck2009fast}.

Notice that the heat kernel parameter $\tau$ becomes just a scale for values in $\boldsymbol L$, with eg. $\tau * \boldsymbol L = (2\tau) *(0.5\boldsymbol L)$. Unfortunately, that means that without any prior knowledge on values in $\boldsymbol L$, it is impossible to uniquely determine the value of $\tau$. We still keep $\tau$ in the formulation, as it is easily determined when the norm or size of values in $\boldsymbol L$ is known a priori. More realistically, its scale is uniquely determined when the heat diffusion process is observed in different moments (i.e. for different $\tau$ values), but on the same set of graphs. In those cases $\tau$ proves to be very important, and apart from significantly changing signal values, it highly affects the accuracy of graph inference, as will be demonstrated in experiments below.

\section{Experimental results}

In this section, we present experimental results that show the effectiveness of our new inference algorithm. We first evaluate the graph heat mixture model (GHMM) on synthetically generated data, comparing it with alternative methods from the literature. We then turn to real data describing Uber pick-ups in Manhattan, where our method manages to automatically separate data corresponding to different mobility patterns at different times of day.

\subsection{Synthetic results}
We first evaluate the performance of our method for different sizes of the observed signal set and different values of the heat kernel parameter $\tau$. We generate two connected Erdos-Renyi graphs $L_1$ and $L_2$ of size 20, with edge probability $p = 0.7$. The means for each cluster are randomly drawn from $\mu_k \sim N(0,0.1 I )$, and the membership probabilities for each cluster are fixed to $\alpha_k = 0.5$.

We compare our method to the Graph Laplacian Mixture model (GLMM) \cite{maretic2018graph}, which jointly infers signal clusters and the corresponding graphs based on mere smoothness priors. We also compare to a Gaussian mixture model (GMM), where the thresholded inverse covariance (precision) matrices act as graph structures; as well as a method performing $K$-means followed by an established graph learning technique \cite{kalofolias2016learn} on each of the clusters separately (K-means + GL). As all methods show high sensitivity to initialisation, we run each experiment 5 times with different random intialisations of $\boldsymbol \mu, \boldsymbol L$ and $\boldsymbol \alpha$, and select the best performing run for each algorithm. We repeat this experiment 100 times, and present results in terms of clustering NMSE (in \%) and graph inference F-measure.

First, we observe the behaviour of our algorithm with respect to a different number of observed signals $M \in \{50, ..., 600\}$. For each $M$ we generate $M/2$ signals from $C_1$ and $M/2$ from $C_2$. The signals are random instances of Gaussian distributions $x_m \sim N(\mu_k, e^{-2\tau L_k})$, where the mean $\mu_k$, $\tau$ and the graph Laplacian $L_k$ drive the heat diffusion processes. 

\begin{figure}
\centering
\begin{subfigure}{.7\linewidth}
	\centering
	\includegraphics[width=\linewidth]{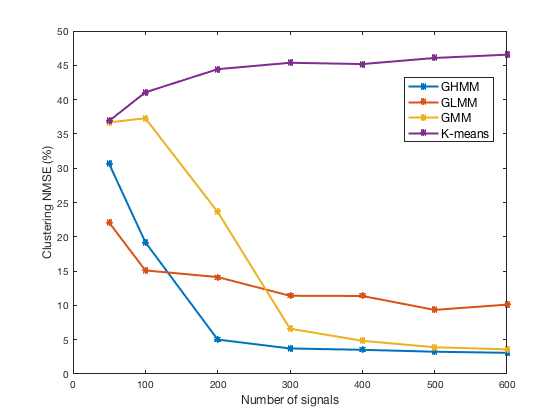}
	\caption{Clustering performance}
\end{subfigure}

\begin{subfigure}{.7\linewidth}
	\centering
	\includegraphics[width=\linewidth]{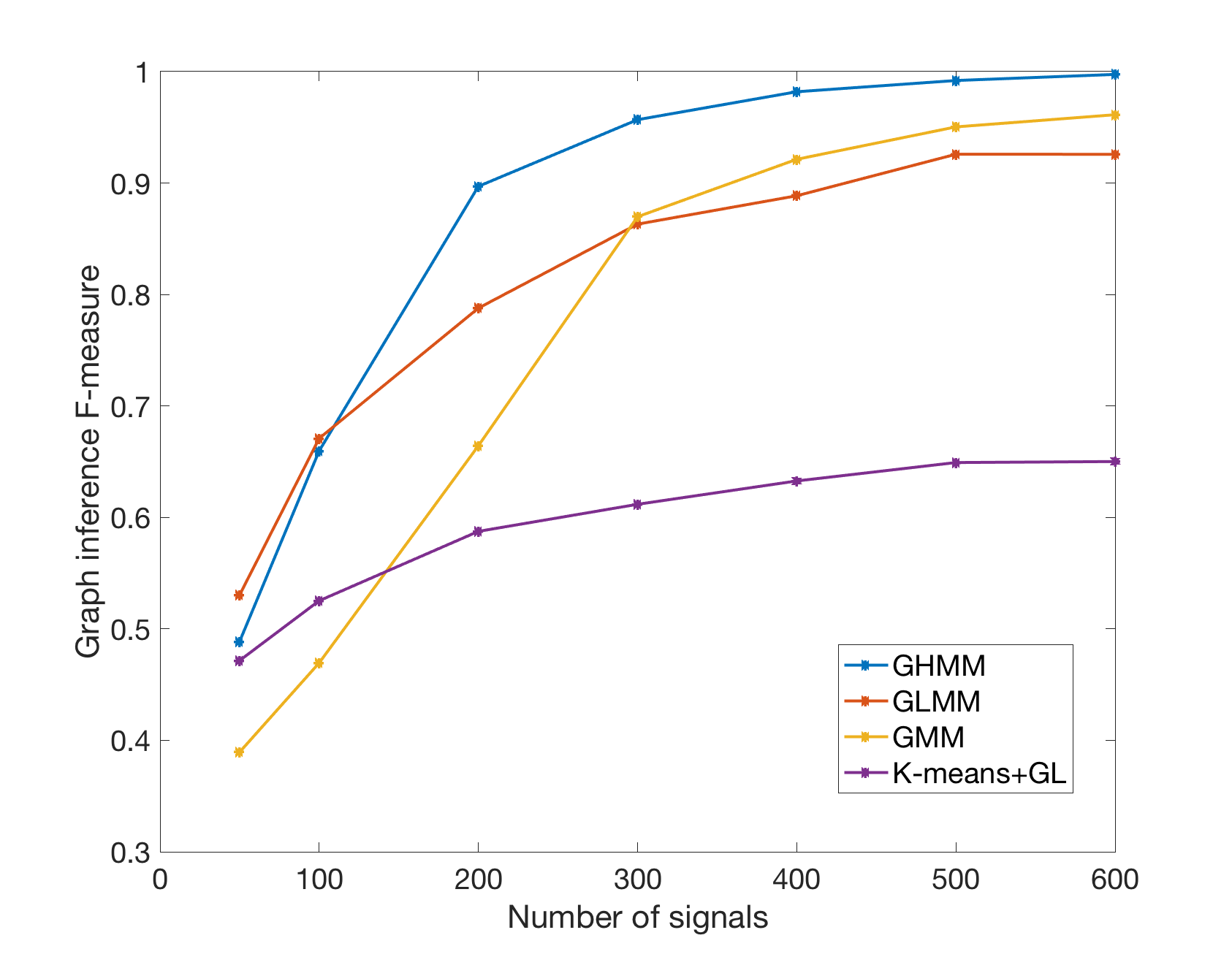}
    \caption{Graph inference}
	\end{subfigure}
	\caption{Performance with respect to different number of available signal observations.}
	\vspace{-1.5em}
	\label{fig:nsig}
\end{figure}

We see in Figure \ref{fig:nsig} that a graph mixture model is very favourable as opposed to separately clustering data and performing graph inference afterwards. The Graph Laplacian mixture model shows slightly better performance for very low sizes of signals sets, due to the fact that the graph inference method in GLMM does not rely on sample covariance matrices, known to be very noisy for small amounts of data. However, in scenarios with larger number of training signals, our new Graph heat mixture model gives the best performance among the methods under comparison, both in terms of clustering error and graph inference F-measure.

\begin{figure}
\centering
\begin{subfigure}{.7\linewidth}
	\centering
	\includegraphics[width=\linewidth]{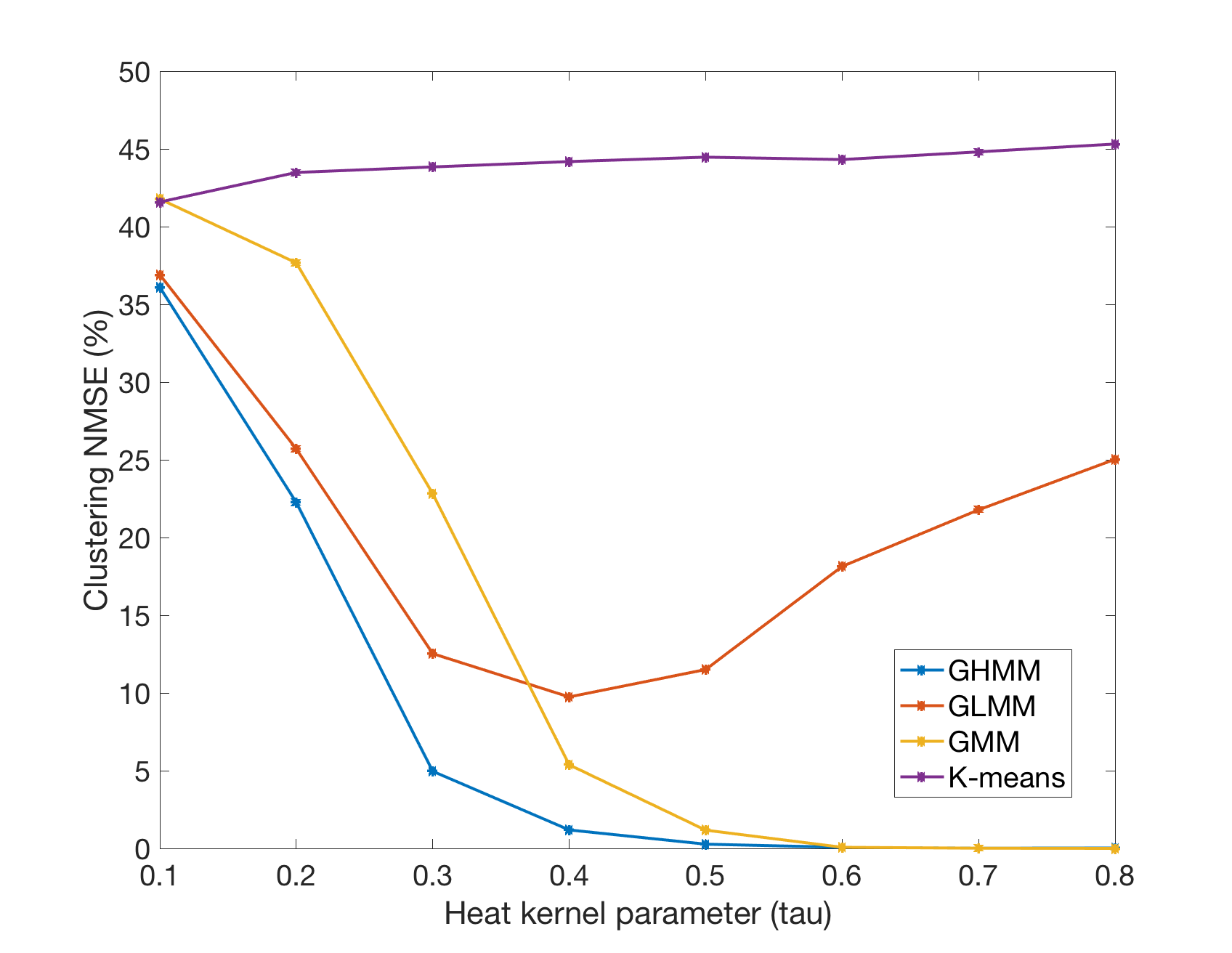}
	\caption{Clustering performance}
	\label{fig:tau_c}
\end{subfigure}

\begin{subfigure}{.7\linewidth}
	\centering
	\includegraphics[width=\linewidth]{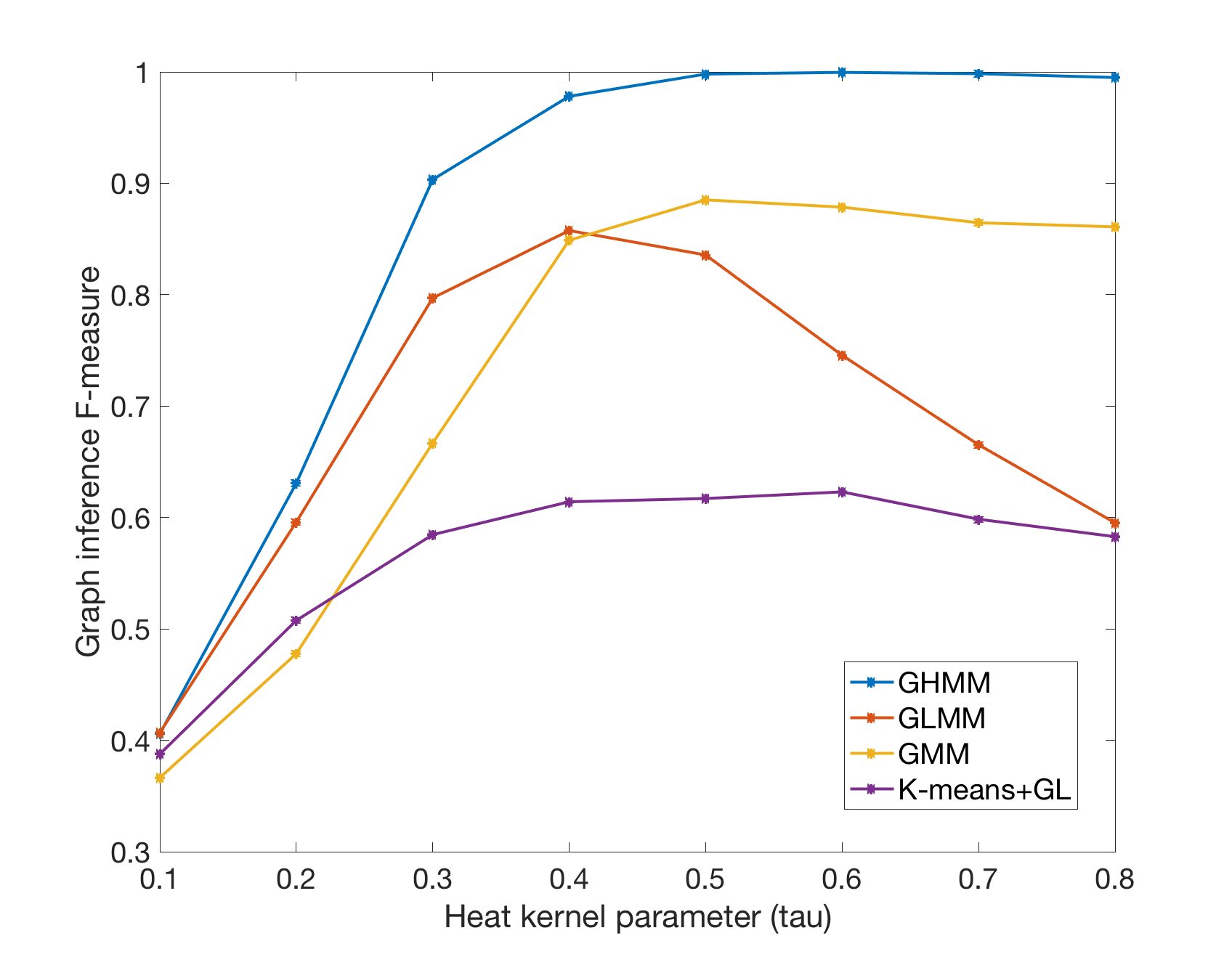}
    \caption{Graph inference}
	\label{fig:tau_f}
	\end{subfigure}
	\caption{Performance with respect to $\tau$.}
	\label{fig:tau}
\end{figure}

We next test the performance of the inference algorithms as a function of the heat parameter $\tau$ that changes between 0.1 and 0.8. The number of signals is fixed to $M = 200$ in this case. Figure \ref{fig:tau} shows that, for very small values of $\tau$, all algorithms have difficulties in recovering the structure as the covariance matrix is close to identity. For large values of $\tau$, the signals that we observe are very smooth. For this reason, the simple smoothness assumption used in GLMM is too weak to successfully separate signals, while our new Graph heat mixture model provides the best performance. The method based on separate clustering and graph learning again performs worst in these experiments. 

\subsection{Uber data}
\begin{figure}[t]
\centering
\begin{subfigure}{.45\linewidth}
	\centering
	\includegraphics[width=\linewidth]{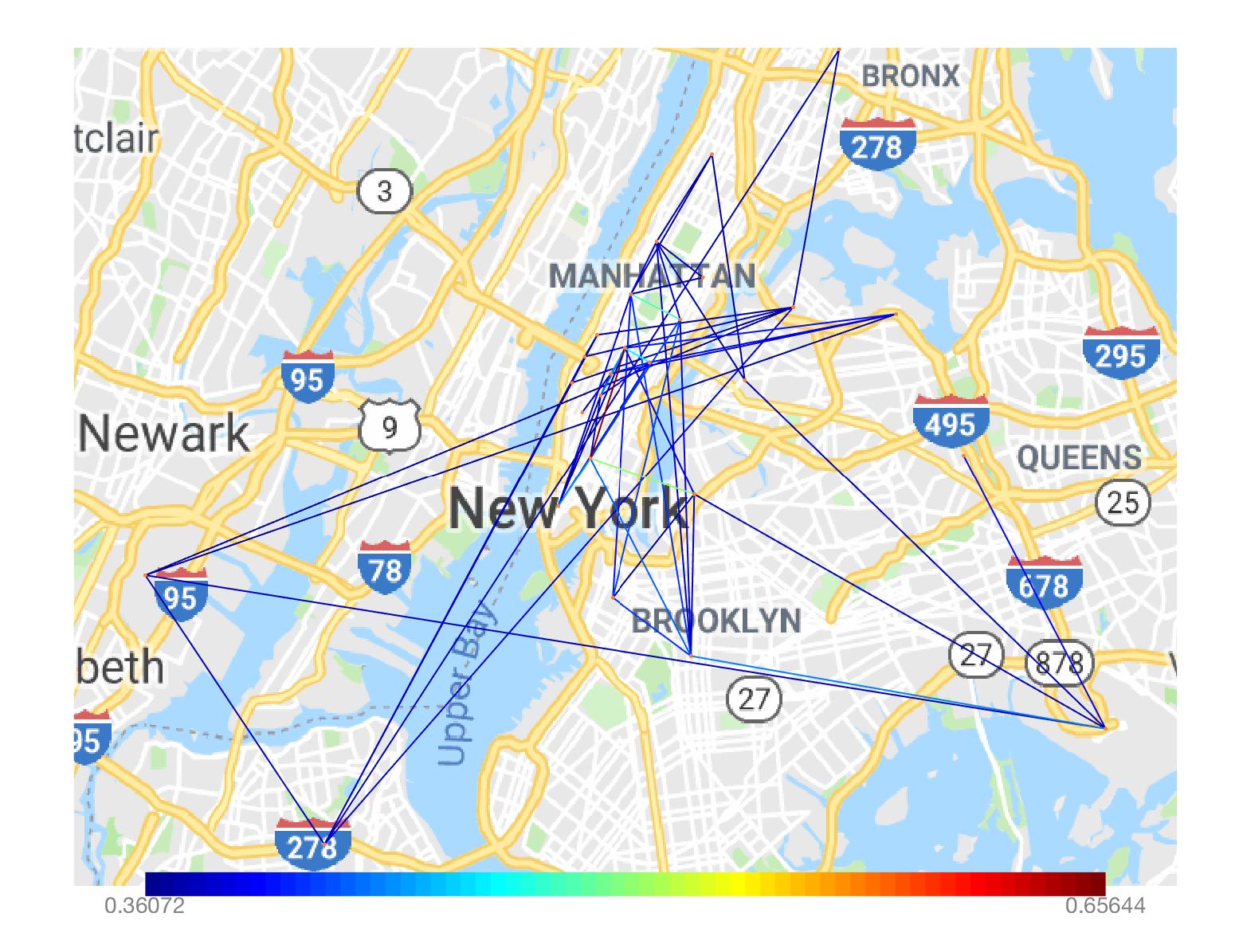}
	\caption{23h - 6h}
\end{subfigure}%
\begin{subfigure}{.45\linewidth}
	\centering
	\includegraphics[width=\linewidth]{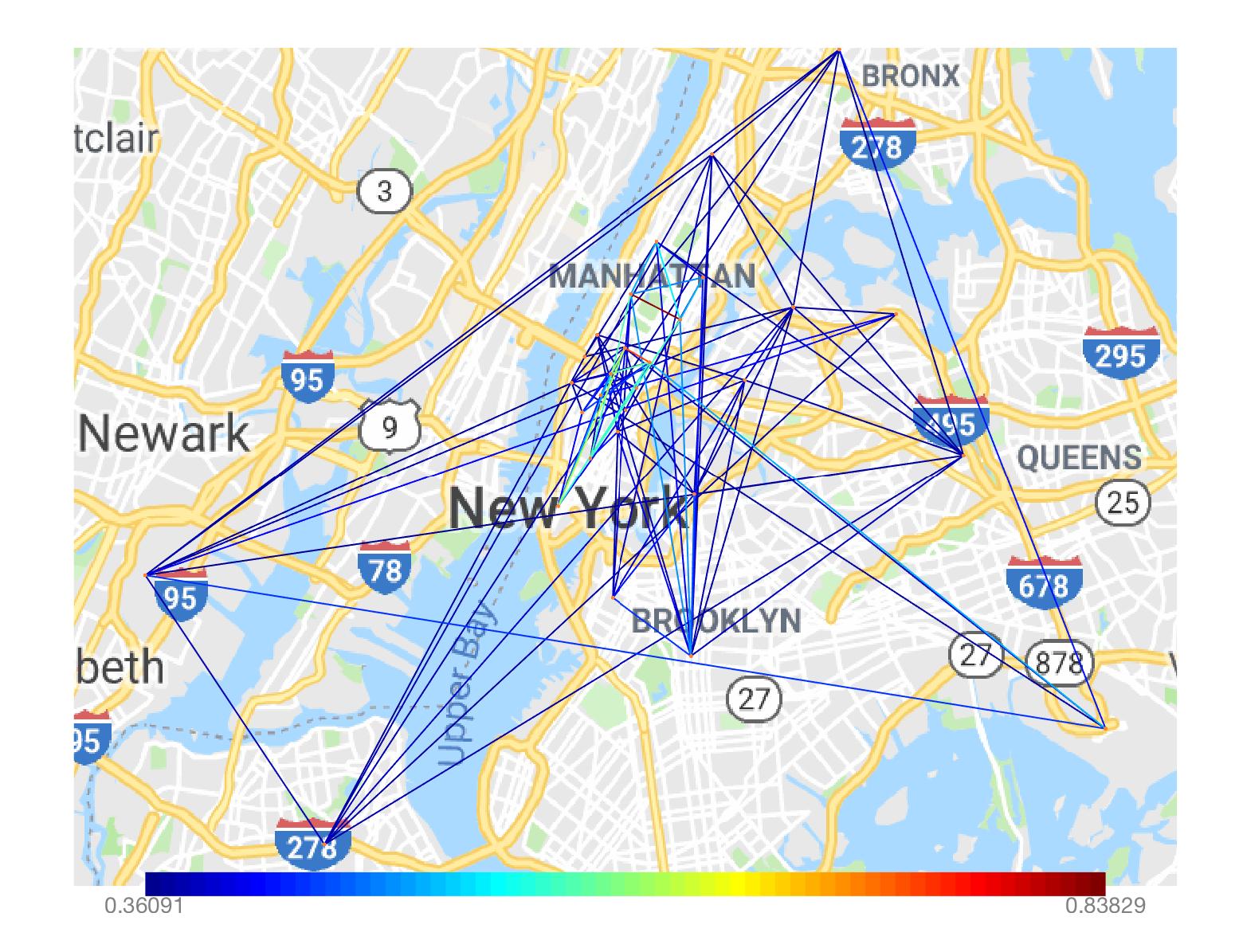}
	\caption{6h - 15h}
\end{subfigure}

\begin{subfigure}{.45\linewidth}
	\centering
	\includegraphics[width=\linewidth]{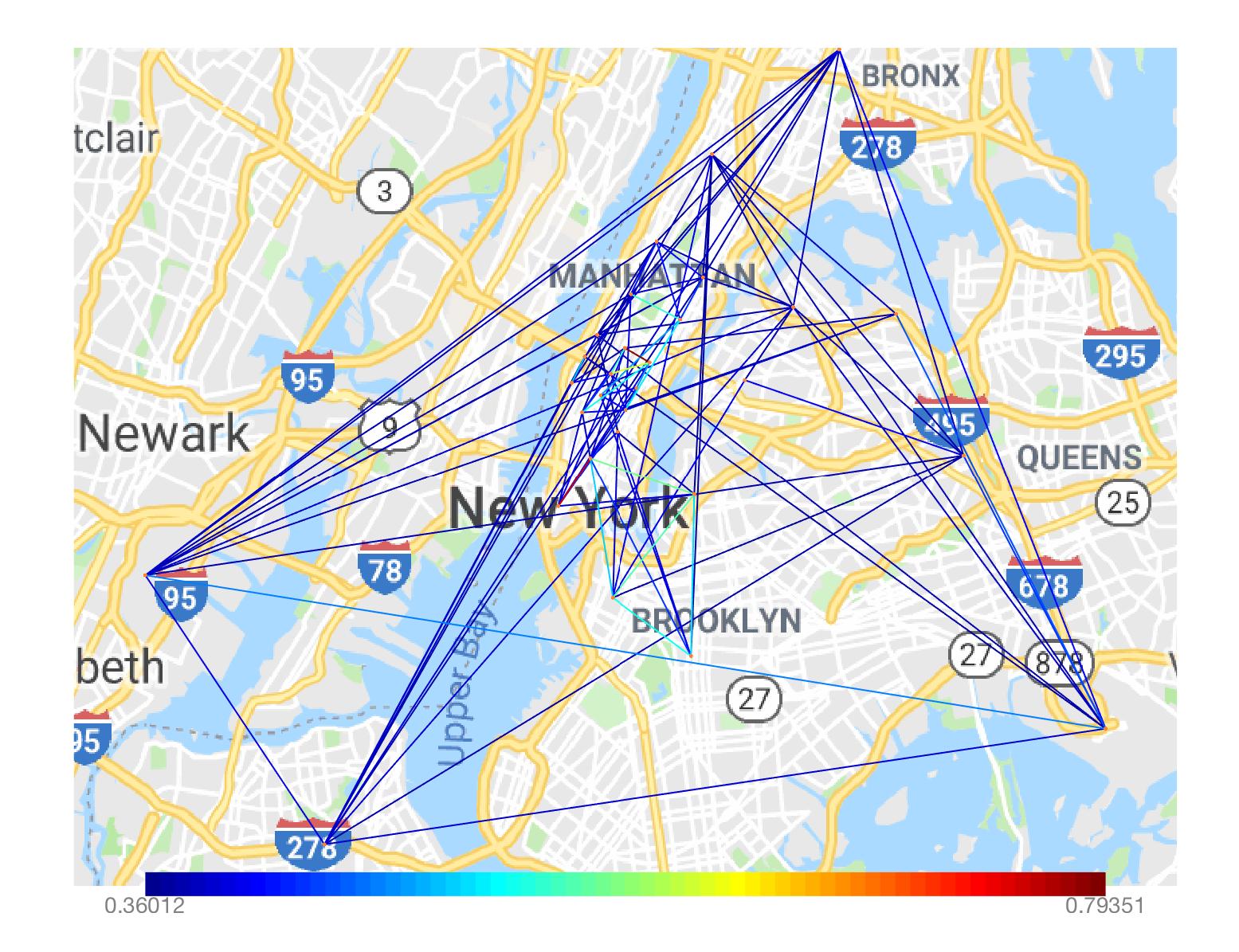}
	\caption{15h - 20h}
\end{subfigure}%
\begin{subfigure}{.45\linewidth}
	\centering
	\includegraphics[width=\linewidth]{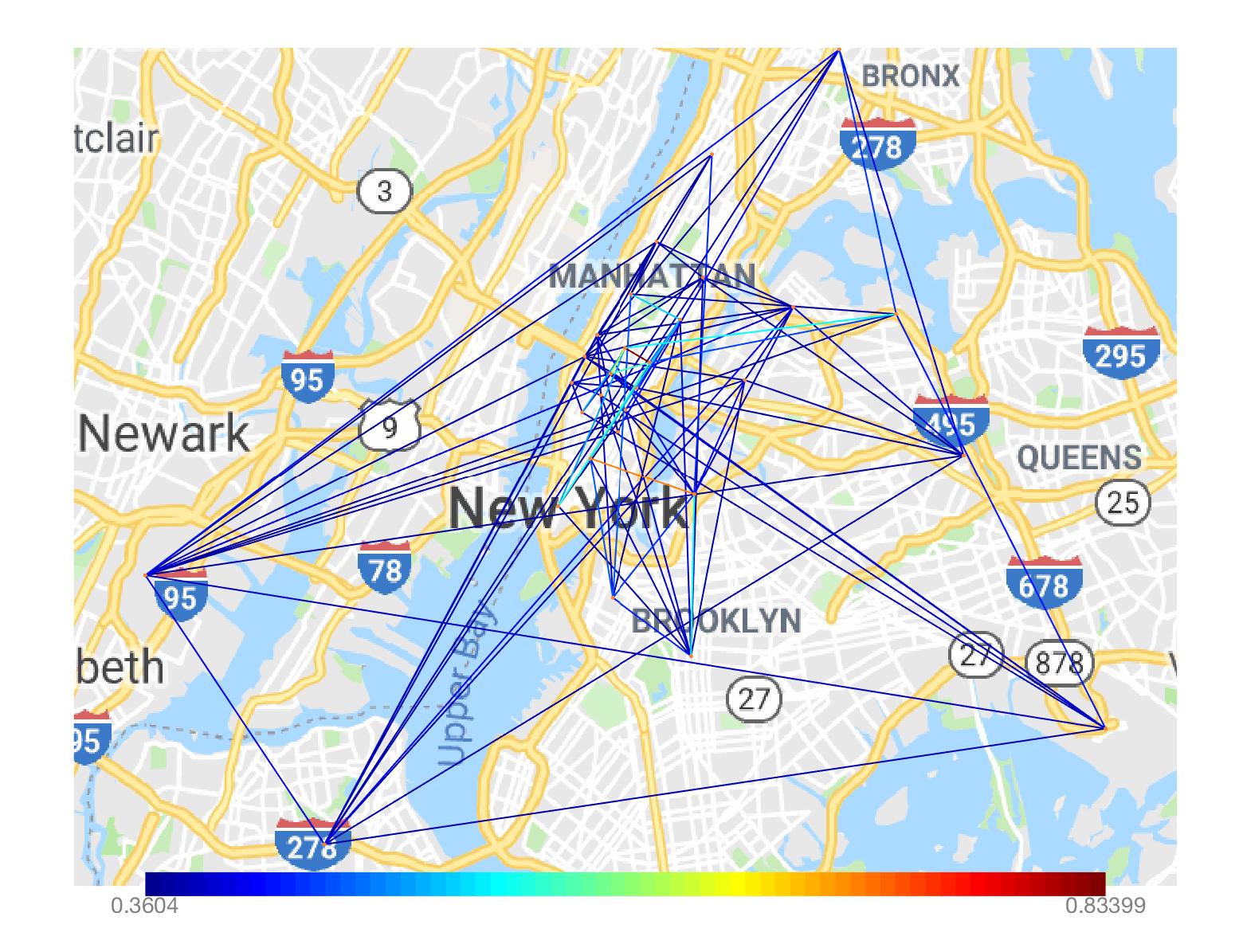}
	\caption{20h - 23h}
\end{subfigure}
\caption{Graphs corresponding to Uber patterns in different times of day.}
\label{fig:uber_graphs}
\end{figure}

We use GHMM to search for patterns in Uber data representing hourly pickups in New York City during the working days of September 2014 \footnote{https://github.com/fivethirtyeight/ubertlc-foil-response}. We divide the city into 29 taxi zones, and treat each zone as a node in our graphs. The signal on these nodes corresponds to the number of Uber pickups in the corresponding zone. We fix the number of clusters to $K = 4$. 
\begin{figure}[t]
	\centering
	\includegraphics[scale = 0.13]{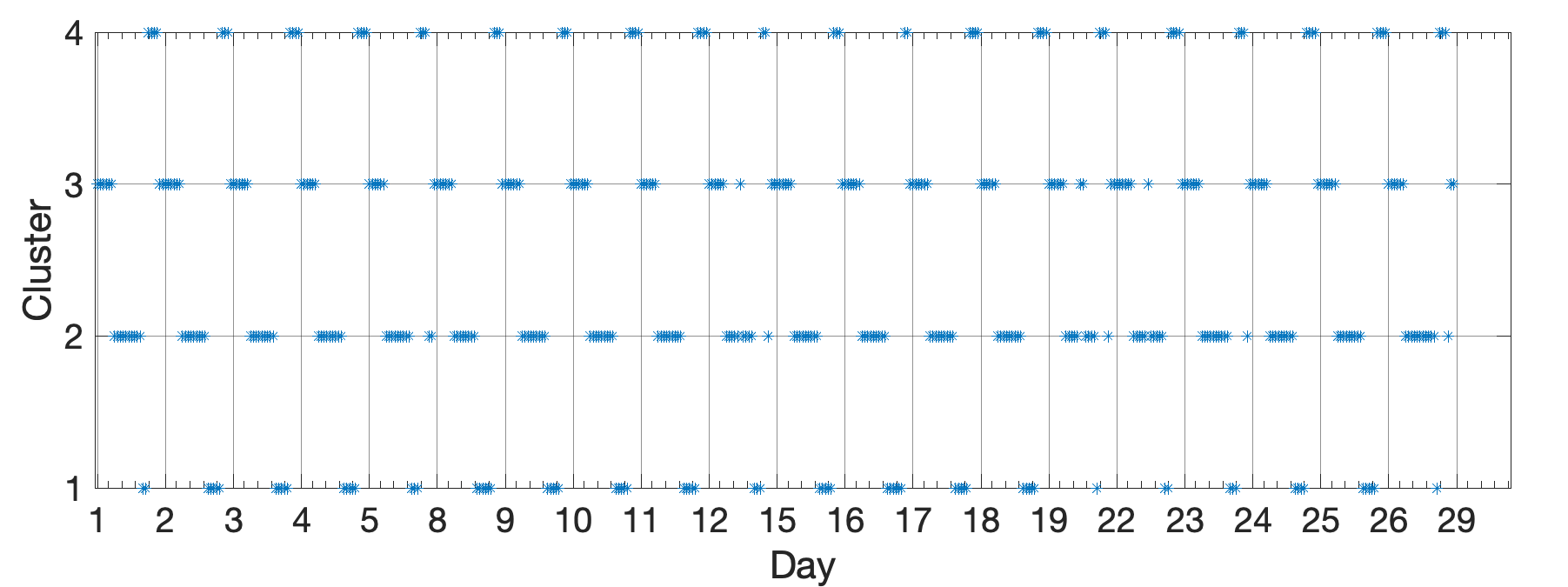}
	\caption{Cluster indexes for Uber hourly signals. Each dot represents one hour in the day, and thin vertical lines represent the beginning of each working day.}\label{fig:ghmm_uber}
	\vspace{-1.5em}
\end{figure}
Figure \ref{fig:ghmm_uber} shows the clustering of hourly Uber signals into 4 different clusters. We can see a slightly noisy periodic pattern, recurring daily. If we inspect the results more carefully, the data in each cluster corresponds to a different period of the day, specifically 23h-6h, 6h-15h, 15h-20h or 20h-23h. In fact, compared to these fixed periods, the clusters inferred with GHMM differ only in a small percentage of observations, with the normalised mean square difference of 7.58\%.
Finally, Figure \ref{fig:uber_graphs} presents different graphs inferred with our method. Each graph shows patterns of a different period in the day. For example, the traffic during nights and early mornings is restricted to the city center and communications with the airports, while direct communication among non-central locations becomes more active later in the day. These different mobility patterns look reasonable with respect to daily people routine in NYC. 

\section{Conclusion}

We have proposed a novel generative model for mixed signals, where each signal is assumed to belong to an unknown cluster and to follow a heat diffusion process on an unknown graph associated to this cluster. In these realistic settings that do not require prior knowledge on signal clusters, we design a new inference method based on a expectation-maximisation algorithm, which can jointly group the signals into clusters and learn their respective graph structures. Experiments on both synthetic and real data show that our new algorithm performs better than alternative inference methods that are based on mere smooth signal priors, or that perform clustering and graph learning separately. 

\bibliographystyle{IEEEbib}
\bibliography{references}

\begin{thebibliography}{10}

\bibitem{dempster1972covariance}
Arthur~P Dempster,
\newblock ``Covariance selection,''
\newblock {\em Biometrics}, pp. 157--175, 1972.

\bibitem{dong2018learning}
Xiaowen Dong, Dorina Thanou, Michael Rabbat, and Pascal Frossard,
\newblock ``Learning graphs from data: A signal representation perspective,''
\newblock {\em arXiv preprint arXiv:1806.00848}, 2018.

\bibitem{Dong14}
X.~Dong, D.~Thanou, P.~Frossard, and P.~Vandergheynst,
\newblock ``Learning laplacian matrix in smooth graph signal representations,''
\newblock {\em IEEE Transactions on Signal Processing}, vol. 64, no. 23, pp.
  6160--6173, 2016.

\bibitem{kalofolias2016learn}
Vassilis Kalofolias,
\newblock ``How to learn a graph from smooth signals,''
\newblock in {\em Artificial Intelligence and Statistics}, 2016, pp. 920--929.

\bibitem{egilmez2017graph}
Hilmi~E Egilmez, Eduardo Pavez, and Antonio Ortega,
\newblock ``Graph learning from data under laplacian and structural
  constraints,''
\newblock {\em IEEE Journal of Selected Topics in Signal Processing}, vol. 11,
  no. 6, pp. 825--841, 2017.

\bibitem{ma2008mining}
Hao Ma, Haixuan Yang, Michael~R Lyu, and Irwin King,
\newblock ``Mining social networks using heat diffusion processes for marketing
  candidates selection,''
\newblock in {\em Proceedings of the 17th ACM conference on Information and
  knowledge management}. ACM, 2008, pp. 233--242.

\bibitem{thanou2017learning}
Dorina Thanou, Xiaowen Dong, Daniel Kressner, and Pascal Frossard,
\newblock ``Learning heat diffusion graphs,''
\newblock {\em IEEE Transactions on Signal and Information Processing over
  Networks}, vol. 3, no. 3, pp. 484--499, 2017.

\bibitem{vlaski2018online}
Stefan Vlaski, Hermina~P Mareti{\'c}, Roula Nassif, Pascal Frossard, and Ali~H
  Sayed,
\newblock ``Online graph learning from sequential data,''
\newblock in {\em 2018 IEEE Data Science Workshop (DSW)}. IEEE, 2018, pp.
  190--194.

\bibitem{egilmez2018graph}
Hilmi~E Egilmez, Eduardo Pavez, and Antonio Ortega,
\newblock ``Graph learning from filtered signals: Graph system and diffusion
  kernel identification,''
\newblock {\em arXiv preprint arXiv:1803.02553}, 2018.

\bibitem{segarra2017network}
Santiago Segarra, Antonio~G Marques, Gonzalo Mateos, and Alejandro Ribeiro,
\newblock ``Network topology inference from spectral templates,''
\newblock {\em IEEE Transactions on Signal and Information Processing over
  Networks}, vol. 3, no. 3, pp. 467--483, 2017.

\bibitem{maretic2017graph}
Hermina~Petric Maretic, Dorina Thanou, and Pascal Frossard,
\newblock ``Graph learning under sparsity priors,''
\newblock in {\em Acoustics, Speech and Signal Processing (ICASSP), 2017 IEEE
  International Conference on}. Ieee, 2017, pp. 6523--6527.

\bibitem{kalofolias2017learning}
Vassilis Kalofolias, Andreas Loukas, Dorina Thanou, and Pascal Frossard,
\newblock ``Learning time varying graphs,''
\newblock in {\em Acoustics, Speech and Signal Processing (ICASSP), 2017 IEEE
  International Conference on}. IEEE, 2017, pp. 2826--2830.

\bibitem{segarra2017joint}
Santiago Segarra, Yuhao Wangt, Caroline Uhler, and Antonio~G Marques,
\newblock ``Joint inference of networks from stationary graph signals,''
\newblock in {\em Signals, Systems, and Computers, 2017 51st Asilomar
  Conference on}. IEEE, 2017, pp. 975--979.

\bibitem{maretic2018graph}
Hermina~Petric Maretic and Pascal Frossard,
\newblock ``Graph laplacian mixture model,''
\newblock {\em arXiv preprint arXiv:1810.10053}, 2018.

\bibitem{beck2009fast}
Amir Beck and Marc Teboulle,
\newblock ``A fast iterative shrinkage-thresholding algorithm for linear
  inverse problems,''
\newblock {\em SIAM journal on imaging sciences}, vol. 2, no. 1, pp. 183--202,
  2009.

\bibitem{chung2007heat}
Fan Chung,
\newblock ``The heat kernel as the pagerank of a graph,''
\newblock {\em Proceedings of the National Academy of Sciences}, vol. 104, no.
  50, pp. 19735--19740, 2007.

\end{thebibliography}

\end{document}